%% file: ms.tex
\pdfoutput=1 
\documentclass[10pt,twocolumn,letterpaper]{article}
\usepackage{authblk}
\usepackage{iccv}
\usepackage{times}
\usepackage{epsfig}
\usepackage{graphicx}
\usepackage{amsmath}
\usepackage{amssymb}
\usepackage{multirow}
\usepackage[table,xcdraw]{xcolor}
\usepackage[accsupp]{axessibility}  
\newcommand{\chk}{\checkmark}
\newcommand{\X}{$\times$}

\usepackage{enumitem}
\usepackage{bm}
\usepackage{subcaption}
\usepackage{varwidth}
\usepackage{mdframed}

\usepackage[breaklinks=true,bookmarks=false]{hyperref}

\iccvfinalcopy 

\def\iccvPaperID{3560} 

\ificcvfinal\fi

\begin{document}


\title{SaccadeCam: Adaptive Visual Attention for Monocular Depth Sensing \vspace{-5mm}}
\author[]{Brevin Tilmon  \hspace{.6cm}  Sanjeev J. Koppal \vspace{-1ex}}
\affil[]{University of Florida}




\maketitle
\ificcvfinal\thispagestyle{empty}\fi

\begin{abstract}
Most monocular depth sensing methods use conventionally captured images that are created without considering scene content. In contrast, animal eyes have fast mechanical motions, called \emph{saccades}, that control how the scene is imaged by the fovea, where resolution is highest. In this paper, we present the SaccadeCam framework for adaptively distributing resolution onto regions of interest in the scene. Our algorithm for adaptive resolution is a self-supervised network and we demonstrate results for end-to-end learning for monocular depth estimation. We also show preliminary results with a real SaccadeCam hardware prototype. 
\end{abstract}

\input{sections/intro}

\input{sections/relatedwork}

\input{sections/background}

\input{sections/experiments}

\input{sections/hardware}

\section{Discussion and Limitations}

In this paper we provide a new framework, SaccadeCam, for leveraging visual attention during image formation. Our key idea is to adaptively distribute resolution onto the scene, to improve depth sensing, demonstrating that our framework can perform better than equiangular distribution of pixels. We now discuss some limitations that we would like to improve in future work:

\noindent \emph{Real-time demonstrations: } Our current hardware prototype allows for on-device end-to-end learning at nearly 5 Hz. We want to demonstrate dynamic scenes results soon with faster hardware.

\noindent \emph{Deformable attention masks: }Our setup and theory already allow deformable attention masks, and we wish to use a liquid lens to demonstrate this.

\noindent \emph{Beyond depth estimation: }The differentiable and modular nature of the SaccadeCam framework encourages integrating SaccadeCam into other existing vision applications such as semantic segmentation or pedestrian detection. \\

\noindent \textbf{Acknowledgements:} The authors thank the following funding agencies for partial support: Office of Naval Research through N00014-18-1-2663 and National Science Foundation through NSF CAREER 1942444 and NSF 1909192.

{\small
\bibliographystyle{ieee_fullname}
\bibliography{ms}
}

\end{document}

%% file: sections/intro.tex
\section{Introduction}
\label{introduction}

\indent Deep depth estimation from a single view has been effective at demonstrating the rich geometric cues available in an image \cite{saxena2008make3d,ranftl2016dense,liu2014discrete,silberman2012indoor,chen2016single}. Additionally, these results are improved by using other cues, such as sparse LIDAR or stereo measurements \cite{uhrig2017sparsity,zhang2018deep,liu2019neural,adaptivelidarstanford}. Our key idea is to notice that most previous monocular approaches assume a nearly equal distribution of sensor pixels across the camera's field-of-view (FOV). In contrast, animal eyes distribute resolution unevenly using fast, mechanical motion, or \emph{saccades}, that change where the eye's fovea views the scene with high acuity. In this paper, we present \emph{SaccadeCam}, a new algorithmic and hardware framework for visual attention control that automatically distributes resolution onto a scene to improve monocular depth estimation. 

\subsection{Why Leverage Attention for Depth Sensing?}

Many methods seek to replicate the biological advantages of attention, such as computational efficiency. However, most efforts apply attention within network training and testing, \emph{after} images have been captured~\cite{ramachandran2019stand, vaswani2017attention, li2018hierarchical,wu2020generative,johnston2020self}. Our framework complements existing attention-based learning, since SaccadeCam leverages visual attention to distribute resolution \emph{during} image capture, and deep attention mechanisms can still be applied after the capture of a SaccadeCam image. Since SaccadeCam can leverage attention during image capture, it can extract novel efficiencies, particularly for bandwidth of image data. The potential for bandwidth reduction is important --- Marr observed that to have foveal resolution everywhere \emph{``...would be wasteful, unnecessary and in violation of our own experience as perceivers..."} \cite{marr1982vision}. SaccadeCam extracts the biological bandwidth advantages of attention, which impacts platforms that need perception within strict budgetary constraints, such as small robots and long-range drones. We show SaccadeCam results for distributing visual attention (using the proxy of image resolution) to improve depth estimation. \textbf{Our contributions are:}

\vspace{-1.5ex}

\begin{figure}[t]
    \centering
    \includegraphics[width=\linewidth]{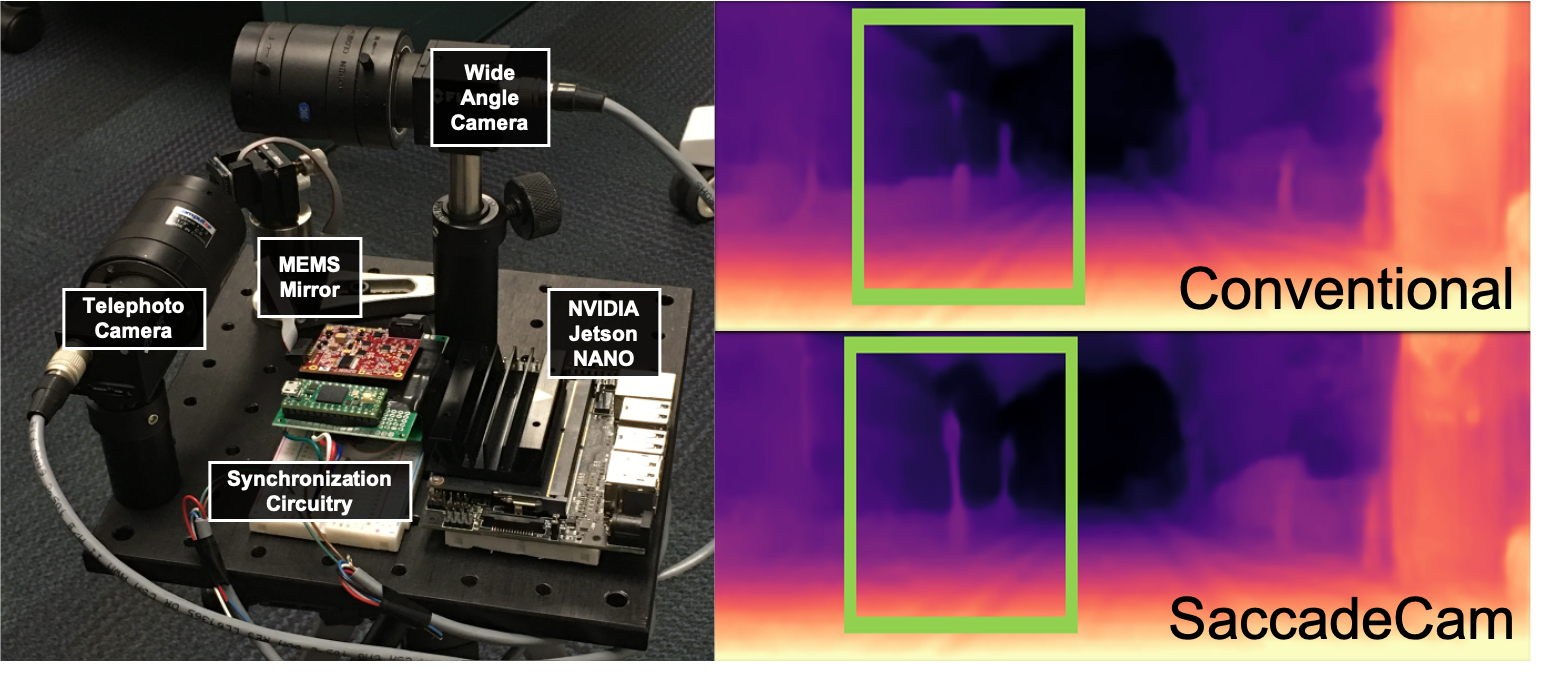}
    \caption{Our method learns to distribute resolution onto regions that improve self-supervised monocular depth estimation while using the same number of pixels as conventional equiangular cameras.}
    \label{fig:teaser}
\end{figure}

\begin{itemize}
    \item We define a new problem of distributing image resolution under a fixed camera bandwidth around the scene with the goal of succeeding at depth estimation (Sect. \ref{sec:background} and Table \ref{tab:oracle_table}). 

\vspace{-1.5ex}

    \item We design an end-to-end network that controls resolution distribution, showing that SaccadeCam images outperform conventional distribution of resolution and can detect important objects for robot navigation, such as poles, signs and distant vehicles (Sect. \ref{sec:algo}, Table \ref{tab:abl1} and Fig. \ref{fig:results}, Sect. \ref{sec:hardware} Fig. \ref{fig:hardware_results}). 
    
    \vspace{-1.5ex}
   
    \item We validate our method on a real hardware prototype that images multiple fovea per frame. We also present a generalized selection algorithm to extract discrete fovea from the attention mask. (Sect. \ref{sec:hardware}).  
\end{itemize}

\input{sections/table1}

%% file: sections/table1.tex
{\begin{table*}
\centering
\resizebox{\linewidth}{!}{%
\begin{tabular}{|c|c|c|c|c|c|}
\hline                    
Method (with few examples) & Adaptive & Test Input & Depth Recovery & Attention \emph{during} image capture & Self/Semi/Guided  \\
\hline                  
Deep Attention Mechanisms \cite{vaswani2017attention,wang2020object,johnston2020self} & \chk  & Mono/Mono+X  & \chk & \X & All  \\
Compressive Imaging \cite{cs3}  & \X & Mono/Mono+X & \chk & \X & All \\
Monocular Depth Estimation \cite{saxena2008make3d,guizilini20203d} & \X   & Mono & \chk & \X & All  \\
Monocular Guided Upsampling \cite{chen2018estimating,garg2019learning} & \X  & Mono+X & \chk & \X  & Semi/Guided \\
Adaptive Guided Upsampling \cite{bartels2019agile,adaptivelidarstanford} & \chk   & Mono+X & \chk & \X & Guided \\
End-to-end Optics \cite{chang2019deep} & \X & Mono & \chk & \X & Guided  \\
Learned Zoom \cite{zhang2019zoom} & \X & Mono & \X & \X & Guided \\
Adaptive Zoom \cite{Uzkent_2020_CVPR} & \chk & Mono & \X & \X & Self \\
\hline
\hline
SaccadeCam (\textbf{Ours}) & \chk & Mono & \chk & \chk & Self  \\ 
\hline
\end{tabular}
}
\caption{SaccadeCam Framework vs. Other Alternatives: To our knowledge, ours is the only work that provides adaptive, monocular depth estimation by manipulating attention inside the camera, during image capture, while being self-supervised.} \label{fig:tabqual}
\end{table*}
}

%% file: sections/relatedwork.tex
\subsection{Related Work}
\label{related_work}

Saccades, attention and related ideas have been studied in robotics and active vision for many years~\cite{aloimonos1988active,bajcsy1988active,frintrop2008attentional,oberlin2017time,dansereau2011plenoptic,frintrop2010computational,bruce2007attention}. In addition, foveal designs to enable high-quality imaging are also common \cite{okumura2011high,hua2008dual,nakao2001panoramic,darrell1996active}. Our SaccadeCam framework is different in three important ways. First, we explore rich distribution of resolution with \emph{multiple fovea}, which has never been demonstrated before for depth estimation. Second, we apply end-to-end learning to find where to place fovea in a scene to estimate monocular depth with non-uniform spatial resolution. Finally, we demonstrate a working SaccadeCam with a microelectromechanical (MEMS) mirror that is \emph{directly controlled by our trained networks}. We now discuss specific groups of related work, summarized in Table \ref{fig:tabqual}. 

\textbf{Attention in Deep Learning: }Attention in deep learning typically involves learning the parameters of transformations of internal weights, so that the network can differentiably focus on specific regions. Recurrent attention networks, spatial transformer networks and Gaussian attention networks all learn such transformations \cite{kosiorek2017hierarchical, kahou2016ratm, gregor2015draw, jaderberg2016spatial}. \cite{ozcinar} show how to optimally select viewing tiles within a FOV for efficient video streaming in VR headsets. There are also approaches that use reinforcement learning for attention when a differentiable attention model is not available \cite{wang2018costaware, Uzkent_2020_CVPR, Uzkent_2020_WACV}. For example, in \cite{Uzkent_2020_CVPR}, the goal is to select from a small, fixed number of high-resolution patches to obtain better classification accuracy. In contrast, in our method, patches can be placed anywhere in the FOV, and SaccadeCam controls where patches are placed for depth estimation. In this sense, we take the goals of deep attention mechanisms \emph{inside the camera}, changing how image resolution is distributed under a fixed camera bandwidth. 

\textbf{Monocular and Guided Depth Completion:} Monocular depth methods have been very successful \cite{saxena2008make3d,ranftl2016dense,liu2014discrete,silberman2012indoor,chen2016single}. A variety of improvements on these methods by applying a ``mono+X" strategy have been proposed~\cite{battrawy2019lidar,chen2018estimating,mal2018sparse,lu2015sparse,uhrig2017sparsity,riegler2016atgv,hui16msgnet} with an available benchmark on the KITTI dataset \cite{uhrig2017sparsity}. Upsampling has been shown with sparse depth \cite{van2019sparse}, single-photon imagers~\cite{adaptivelidarstanford} and flash lidar \cite{gruber2019gated2depth}. SaccadeCam can be seen a first step towards physical instantiations of recent depth estimation methods that seek to self-improve imperfect measurements~\cite{uhrig2017sparsity,zhang2018deep,liu2019neural,adaptivelidarstanford,pittaluga2020towards}. In contrast to these other approaches, our method is a fully passive approach that adaptively distributes resolution to enable successful monocular estimation, see Table \ref{fig:tabqual}. 

\input{ablation/oracle_final} 

\textbf{Foveated Rendering in VR/AR:} Foveation based on eye tracking is used to bypass rendering entire resolution frames in VR/AR headsets \cite{foveated_3D, DeepFovea}. \cite{DeepFovea} proposed a GAN reconstruction network that is able to take roughly 10\% of an image as input and reconstruct a plausible foveated video. Rather than generating compelling viewing, we are interested in foveated imagery for depth estimation. 

\textbf{Compressive Sensing for Vision:} Compressive signal processing uses coded optics during capture for applications such as classification~\cite{cs1,cs2,cs3}. Compressive sensing optimizes bandwidth at the cost of computing (such as L1-optimization), after image capture, to decode the measurements. Our approach is about emphasizing scene areas with new measurements during image capture, reducing bandwidth without extra computing. 

\textbf{Adaptive Imaging for Vision:} End-to-end learning inside the camera has impacted many applications in computational cameras and computer vision. These include learning optimal structured light patterns \cite{baek2020polka}, learning optimal lens parameters for monocular depth estimation \cite{chang2019deep} and HDR imaging \cite{metzler2019deep}, and learning optimal sensor designs \cite{chakrabarti2016learning}. SaccadeCam is different in that the optics are not fixed but foveate, enabling active, adaptive changes in imaging inside the camera. This is also what separates us from previous work that does not use learning to decide where to distribute resolution \cite{tilmon2020foveacam}. In this sense, our work is similar to adaptive LIDAR work~\cite{Raaj_2021_CVPR, liu2019neural,adaptivelidarstanford,pittaluga2020towards}, but instead we seek to control monocular resolution for depth sensing.

%% file: ablation/oracle_final.tex
\begin{table*}[]
\centering
\resizebox{\textwidth}{!}{\begin{tabular}{|l|l|c|c|c|c|c|c|c|}
\hline
 &  & \cellcolor[HTML]{E18E96}Abs Rel & \cellcolor[HTML]{E18E96}Sq Rel & \cellcolor[HTML]{E18E96}RMSE & \cellcolor[HTML]{E18E96}\begin{tabular}[c]{@{}c@{}}RMSE\\ log\end{tabular} & \cellcolor[HTML]{73C2FB}$\delta \textless 1.25$ & \cellcolor[HTML]{73C2FB}$\delta \textless 1.25^2$ & \cellcolor[HTML]{73C2FB}$\delta \textless 1.25^3$ \\ \hline
 & Full Resolution (70 pixels/mm) & 0.109 & 0.883 & 4.960 & 0.208 & 0.865 & 0.949 & 0.975 \\ \hline
 & Target resolution (31 pixels/mm) & 0.118 & 0.988 & 5.188 & 0.214 & 0.851 & 0.944 & 0.974 \\ \hline \hline
 & Wide Angle Camera (27 pixels/mm) & 0.119 & 0.991 & 5.238 & 0.216 & 0.846 & 0.943 & 0.974 \\ \cline{2-9} 
 & Photometric Oracle & 0.116 & 0.941 & 5.134 & 0.213 & 0.851 & 0.945 & 0.975 \\ \cline{2-9} 
\multirow{-3}{*}{(a)} & \textbf{True Oracle} & \textbf{0.114} & \textbf{0.853} & \textbf{4.850} & \textbf{0.208} & \textbf{0.857} & \textbf{0.950} & \textbf{0.976} \\ \hline \hline
 & Wide Angle Camera (22 pixels/mm) & 0.121 & 1.005 & 5.275 & 0.219 & 0.840 & 0.939 & 0.973 \\ \cline{2-9} 
 & Photometric Oracle & 0.116 & 0.931 & 5.114 & 0.214 & 0.848 & 0.943 & 0.974 \\ \cline{2-9} 
\multirow{-3}{*}{(b)} & \textbf{True Oracle} & \textbf{0.111} & \textbf{0.850} & \textbf{4.846} & \textbf{0.206} & \textbf{0.863} & \textbf{0.950} & \textbf{0.976} \\ \hline \hline
 & Wide Angle Camera (15 pixels/mm) & 0.128 & 1.067 & 5.507 & 0.228 & 0.824 & 0.934 & 0.971 \\ \cline{2-9} 
 & Photometric Oracle & 0.120 & 0.960 & 5.238 & 0.219 & 0.840 & 0.941 & 0.973 \\ \cline{2-9} 
\multirow{-3}{*}{(c)} & \textbf{True Oracle} & \textbf{0.112} & \textbf{0.847} & \textbf{4.848} & \textbf{0.206} & \textbf{0.866} & \textbf{0.951} & \textbf{0.976} \\ \hline

\end{tabular}}
\caption{We motivate our method with a compelling example from the KITTI dataset \cite{geiger2013vision}. We want to see if increasing resolution in high error regions in the wide angle camera (WAC) lowers the overall depth error compared to the target resolution. We compare a full resolution of $70$ px/mm (conventional KITTI imagery) with a target resolution of $31.30$ px/mm. As expected, full resolution does far better than both target resolution and three different resolution low-res WAC images. Photometric Oracle distributes resolution based on per pixel error between WAC-trained and full resolution-trained depth networks. True Oracle distributes resolution based on the error regions between a WAC-trained depth network and ground truth LIDAR. For both True Oracle and Photometric Oracle, depth from a focused-trained depth network with focused input images is placed in the attention regions and then the error is recalculated for the result. The results show that \emph{distributing resolution} adaptively can provide the best depth. In Sect. \ref{sec:algo} we describe our algorithms to extract this potential of SaccadeCam. 
}
\label{tab:oracle_table}
\end{table*}


%% file: sections/background.tex
\section{Can Adaptive Attention Improve Depth?}
\label{sec:background}








Our hypothesis is that distributing pixels within a camera field-of-view can positively impact monocular depth estimation. This is only possible if models of differing bandwidths perform similarly on smooth consistent regions and perform differently on critical regions. We want to test this hypothesis and build learning mechanisms to distribute these pixels in a self-supervised manner, with no requirement for ground truth labels as recent work has shown \cite{monodepth2}. Given a fixed bandwidth, the reduction of resolution in some areas frees up resolution to place onto critical regions such as pedestrians, signs, cars and foliage. In the next section, we discuss how to decide where to place the resolution  and demonstrate the validity of our hypothesis. Now, we discuss the implications of our approach in Table \ref{tab:oracle_table}.



\begin{figure*}
  \includegraphics[width=\textwidth]{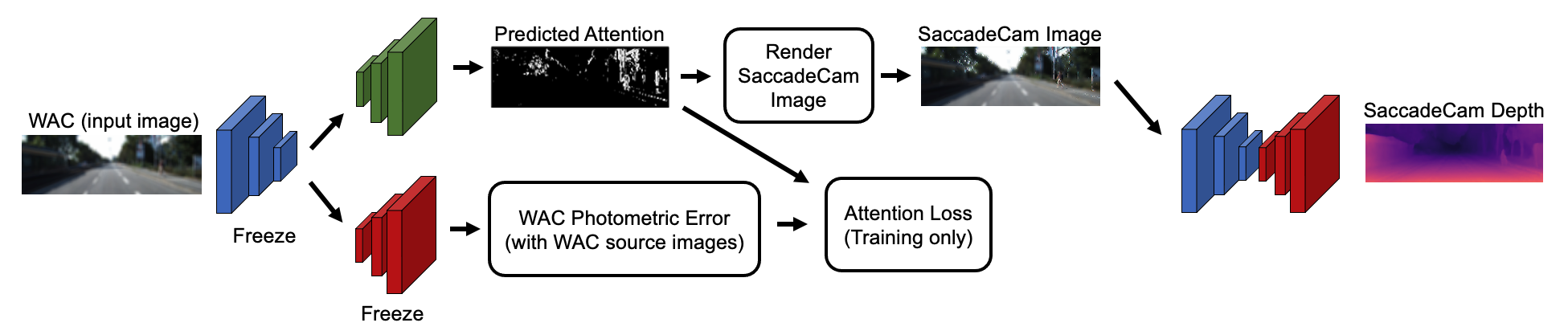}
  \caption{Our Method. We use a self-supervised setup, where the network consists of a single encoder (blue) and two decoders (red and green). Our framework takes an input image and nearby source images during training, and a single input image during testing.}
  \label{fig:method}
\end{figure*}

\subsection{Bandwidth}

Table \ref{tab:oracle_table} has three baselines at different bandwidths. \textbf{We define bandwidth} as the number of angular samples across the FOV, i.e. our notion of bandwidth is identical to angular resolution. Therefore, while for practical reasons we may show images of the same spatial resolution (i.e. pixels in computer memory), they are of very different angular resolution. For all our experiments we use images with camera intrinsics from the KITTI dataset \cite{geiger2013vision}, from which we simulate different camera resolutions. 


We simulate bandwidth by downsampling based on the scaled intrinsic matrix and then upsampling back to original resolution. This simulates a camera that, in practice, would have less resolution bandwidth over the same field of view. The three baselines in Table \ref{tab:oracle_table} are full resolution (70 px/mm bandwidth), target resolution (31.30 px/mm bandwidth) and three low-resolution images that we term as wide-angle camera (WAC) bandwidth in the context of the SaccadeCam hardware in Sect. \ref{sec:hardware}. 

\subsection{Depth from SaccadeCam Images}

In our experiments we use the ground truth color images as the full resolution. The high resolution attention regions in our SaccadeCam images are also at the full resolution. We compare equiangular sampling of the target resolution with SaccadeCam images that have to be at the same bandwidth as the target resolution. \emph{SaccadeCam images are created by fusing high resolution images into attention regions within the low-resolution WAC images.} The WAC resolution and the number of attention regions are constrained by the fact that their sum must equal the target angular resolution. While monocular images with equiangular resolutions have a variety of methods for depth estimation, these cannot be used directly on SaccadeCam images without training or fine tuning. This is because SaccadeCam images have spatially varying resolution, and in Sect. \ref{sec:algo} we discuss how to extract depth from such monocular imagery. Now we discuss the implications of what is possible if such SaccadeCam depth estimation is solved. 

\subsection{Oracles} Our approach is to compare monocular depth estimation of equiangular images with SaccadeCam images, created by unevenly distributed resolution. We design oracle experiments that determine ideal locations to distribute resolution to, and then place focused depth predictions as a perfect color-to-depth mapping in the attention regions. 

For the Photometric Oracle in Table \ref{tab:oracle_table}, the attention regions are computed based on the top $N$ locations of the difference between the WAC depth prediction errors from a fully trained WAC network and full resolution depth prediction errors from a fully trained full resolution network using the method of \cite{monodepth2}. We then replace the WAC depth with focused depth in the attention regions. $N$ is the limit of available pixels left after the target resolution and WAC resolutions are determined from our camera model. We hypothesize that the focused depth errors should be lower than WAC depth errors in high resolution attention regions and similar to WAC depth errors in smooth geometrically consistent regions. 

For the True Oracle in Table \ref{tab:oracle_table}, the attention regions are computed based on the top $N$ locations of the difference between WAC depth and ground truth LIDAR, where $N$ is scaled according to the number of LIDAR samples versus full resolution for fair comparison. We then replace the WAC depth with focused depth in the attention regions. Therefore, if the worst depth estimates of WAC images are replaced by the corresponding depths in the same regions of full resolution images, then, as can be seen by the Table \ref{tab:oracle_table}, depth from SaccadeCam \emph{has the potential} to outperform state-of-the-art. Our oracle experiments support our idea that better resolution can help with depth estimation as suggested in \cite{marr1982vision, monodepth2}.

%% file: sections/experiments.tex



\section{End-to-end Learning for Adaptive Attention}
\label{sec:algo}

In Figure \ref{fig:method} we depict the complete flow for our self-supervised method. Our system consists of one encoder (blue) and two decoders (red and green). Each of these are designed for self-supervised stereo, following the method of \cite{monodepth2}. Our method could easily be integrated with self-supervised monocular training as well, since the pose can be estimated from multiple views of a single camera using a pose network. At test time the flow in Fig. \ref{fig:method} is monocular (single image), but at training time, each network takes a stereo pair. 

\begin{figure*}[t]
\centering
     \includegraphics[width=\linewidth]{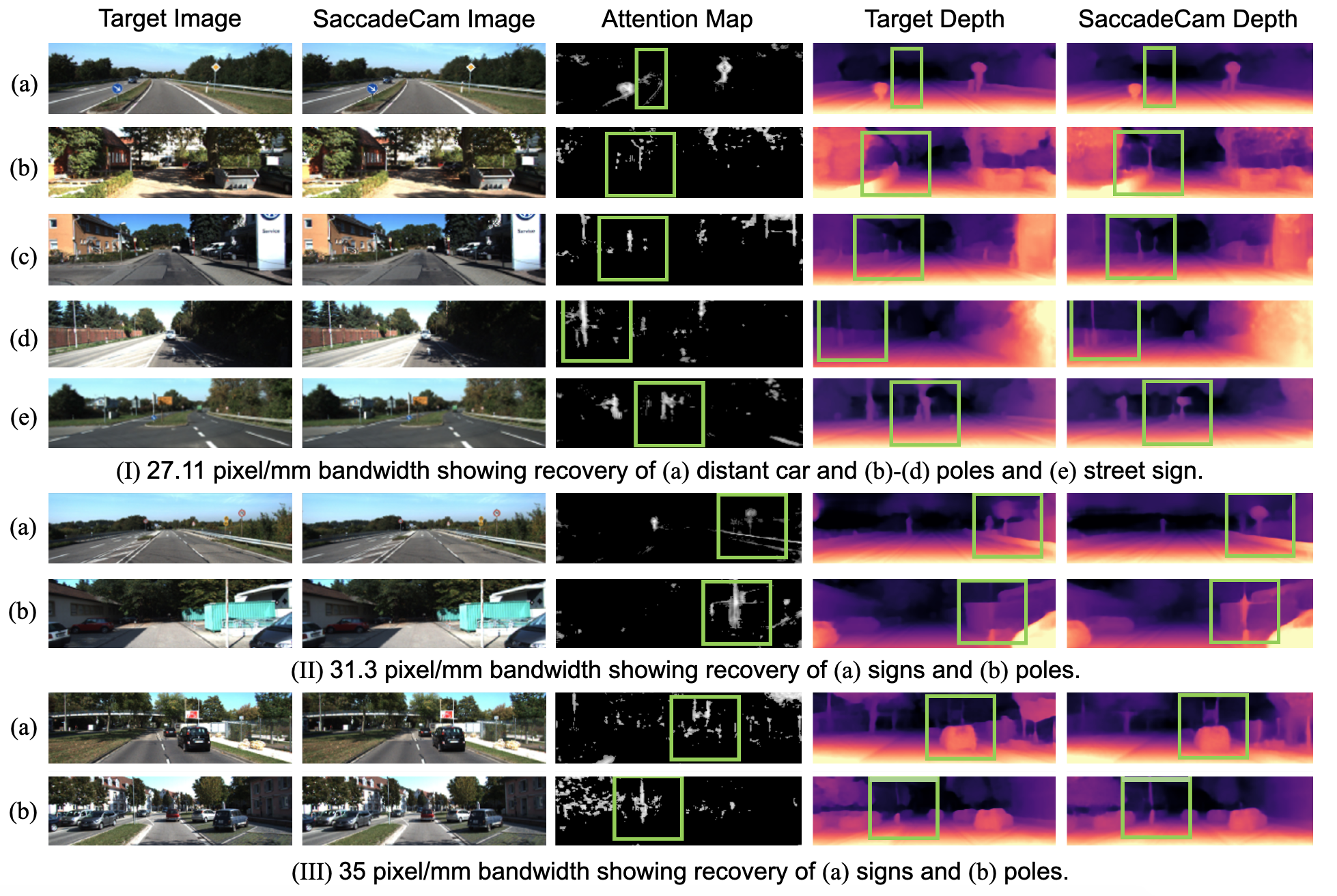}
      \caption{Overview of our KITTI results. In (I-III) we show testing results from our SaccadeCam framework with progressively increasing bandwidth. Our method is particularly good at recovering thin objects such as poles or signs, that can be dangerously ignored by conventional, equiangular sampling of the scene at low resolution.}
      \label{fig:results}
\end{figure*} 

\textbf{Adaptive Attention:} The attention decoder (green in Figure \ref{fig:method}) is trained with a stereo pair of low-resolution, wide angle camera (WAC) images. The attention decoder input is the latent vector of the training depth encoder. The attention decoder then predicts per pixel attention and calculates binary cross entropy loss against the ``true" binary attention mask given by the top photometric error regions calculated from the training depth network. This trains the attention mask towards 1. Our insight is that these error regions should be where additional resolution might make a difference. However, we are not strictly tied to the photometric error, as we will soon see. We then differentiably render a SaccadeCam image using the predicted attention mask, focused image, and WAC image. Here the bandwidth is given by the maximum number of samples that are possible at the highest resolution of the system. The bandwidth is a function of the target resolution and the amount of bandwidth that has already been used up by the WAC image. 

\textbf{SaccadeCam Rendering:} Our SaccadeCam rendering module consists of alpha blending a focused image onto the WAC image using an attention mask as the blend weight. We use this to create SaccadeCam images from either a learned or oracle attention mask \textbf{M}. This allows us to differentiably train our attention network end to end with a downstream monocular network,
\begin{equation}
    \textbf{$I_{SaccadeCam}$} = \textbf{M} \odot (\textbf{$I_{focused}$}) + (1-\textbf{M}) \odot (\textbf{$I_{WAC}$}).
\end{equation} 

\input{ablation/ablation1_final}

\textbf{Depth Network and Attention Regularization:} The last module is the encoder-decoder pair (blue and red) that converts the SaccadeCam image into a depth. When calculating the view synthesis photometric loss \cite{monodepth2}, we compute the loss between the target SaccadeCam image and the synthesized target image that is also foveated with the same attention mask, but with the synthesized focused target image in the attention regions.  The encoder and decoder used in SaccadeCam depth estimation are the same used in obtaining the WAC depth during attention estimates. During attention estimation, the gradients of the depth encoder and decoder pair are frozen. In other words, the encoder and decoder drifts towards monocular SaccadeCam image depth reconstruction, while also regularizing attention estimates. Practically, such a system is more efficient since it shares SaccadeCam features with the attention module and allows for flexible attention beyond WAC photometric errors.

\textbf{Loss Terms:} Our final loss is $L = \mu L_p + \lambda L_s + \alpha L_a$. $L_p$ and $L_s$ follow the view synthesis photometric loss and depth smoothness loss common in monocular depth estimation. We set $\mu = 1$ to avoid masking out fovea regions and $\lambda = 0.001$. $L_a$ is the binary cross entropy loss between the predicted attention and WAC photometric error given by the SaccadeCam depth network. We freeze our depth network and set $\mu = \lambda = 0$, $\alpha = 1$ when training the attention network. We found that the attention decoder learned much quicker than the depth network (roughly 5 epochs for attention compared to roughly 20 epochs for high bandwidth depth). We also found that an attention network trained on a single bandwidth generalizes well across different bandwidths. In an online setting, we hypothesize that infrequently updating or significantly lowering the learning rate of the attention network relative to the depth network would be beneficial.

\section{Experiments}

We implement our network in PyTorch on a single NVIDIA GTX 1080 Ti. Our encoder architecture is a ResNet18 and our decoder architecture is similar to \cite{monodepth2}. All our training was initialized with ImageNet parameters. In Table \ref{tab:abl1}, we show our results over a few different bandwidths. We found our SaccadeCam depth networks finished training earlier than networks trained on equiangular images based on the validation error. We train the depth networks of (a), (b), (c) for 17, 11, 2 epochs respectively and the attention networks of (a), (b), (c) for 5 epochs each. We train all equiangular resolution models for 20 epochs. \emph{Note that not all bandwidths are appropriate for SaccadeCam.} For example, extremely high-resolution images may not benefit from bandwidth optimization, and very low resolution images may result in extreme WAC depth errors. 

\vspace{-.25ex}

We also explored weighting our loss with a weighted binary version of the predicted attention mask based on the observation that high resolution models train longer than low resolution models, this supports giving the high resolution attention region more weighting during training since the periphery is lower resolution. We train the weighted variants of (a), (b), (c) for 7, 14, 1 epochs respectively. Overall the region weighting boosts performance and speeds up training. We found at higher bandwidth SaccadeCam data the region weighting delta must be smaller because, while the periphery is lower resolution than the high resolution attention region, it is still high enough resolution that it needs a stronger weighting to train. We weight the foveal/WAC regions of the photometric error 1.15/0.85, 1.25/0.75, 1.5/0.5 for (a), (b), and (c) respectively in Table \ref{tab:abl1}. 

\begin{figure*}[t]
\centering
     \includegraphics[width=\linewidth]{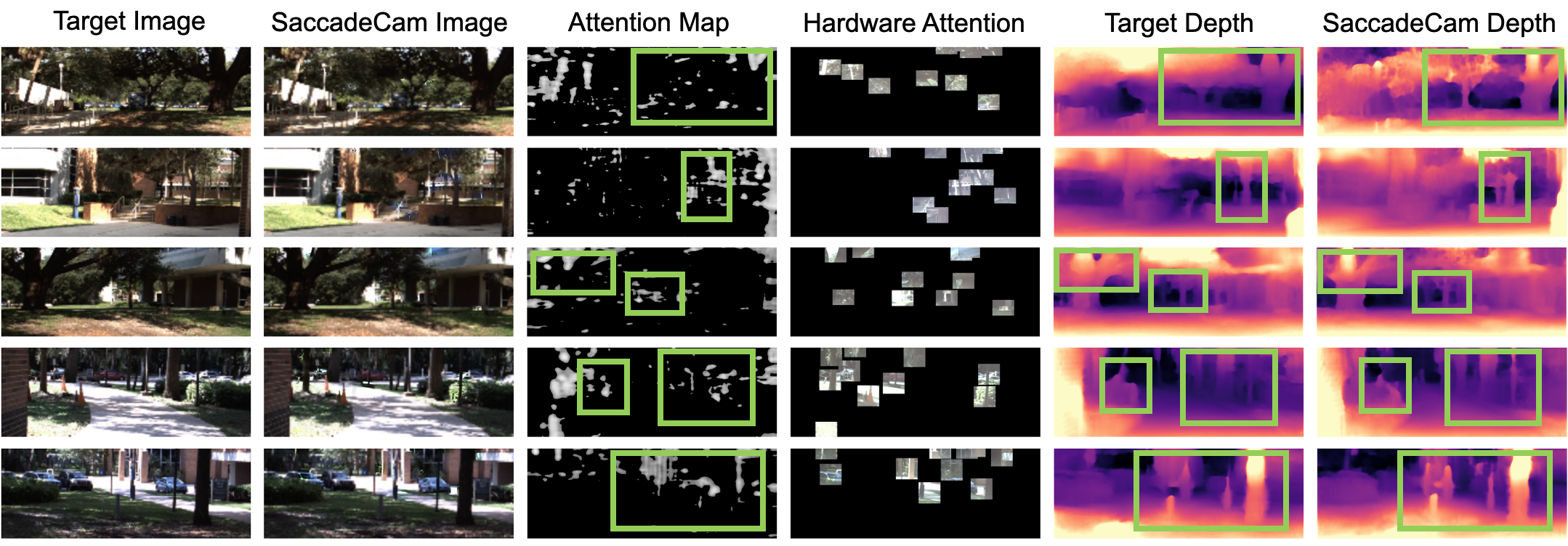}
      \caption{Results for real data captured with our SaccadeCam hardware prototype. Our trained models run on our SaccadeCam hardware prototype to adaptively control the MEMS mirror for approximating the learned attention based on the greedy algorithm. The Hardware Attention column depicts the raw output of the MEMS images that approximate the predicted attention from our models. Our prototype is fast enough to take multiple MEMS images per WAC frame, enabling adaptive resolution distribution to several scene regions on-device at video rate. }
      \label{fig:hardware_results}
\end{figure*} 

\vspace{-.25ex}

We compare our results to monocular self-supervised depth reconstruction at the target resolution. We also compare to a color edge detector as an attention proxy. We found that edges performed well at very low resolutions, but performed poorly at higher resolutions where the fovea must be more intelligently placed to meaningfully impact performance. For our SaccadeCam networks, we first train our depth networks using the WAC photometric error as an attention proxy. We then train the attention network with the same frozen depth network using the WAC photometric error as psuedo ground truth as described in Section \ref{sec:algo}. At test time, we use the learned attention mask. We found $\geq$ 95\% overlap between the predicted attention masks and error regions on average for the test set across bandwidths, which shows the attention masks sufficiently learned to represent the error regions.

Fig. \ref{fig:results} shows visual results from our SaccadeCam models. Our hypothesis holds true in that we perform similar to equiangular models on smooth and geometrically consistent scene regions while outperforming equiangular models on irregular edge-case regions. Notice the SaccadeCam framework allows us to detect road signs, poles, and other distant objects such as cars that the equiangular models cannot detect.

%% file: ablation/ablation1_final.tex
\begin{table*}[]
\centering
\resizebox{\textwidth}{!}{\begin{tabular}{|l|l|l|c|c|c|c|c|c|c|}
\hline
 &  & \multicolumn{1}{c|}{\begin{tabular}[c]{@{}c@{}}Fovea \\ weighting\end{tabular}} & \cellcolor[HTML]{E18E96}Abs Rel & \cellcolor[HTML]{E18E96}Sq Rel & \cellcolor[HTML]{E18E96}RMSE & \cellcolor[HTML]{E18E96}\begin{tabular}[c]{@{}c@{}}RMSE\\ log\end{tabular} & \cellcolor[HTML]{73C2FB}$\delta \textless 1.25$ & \cellcolor[HTML]{73C2FB}$\delta \textless 1.25^2$ & \cellcolor[HTML]{73C2FB}$\delta \textless 1.25^3$ \\ \hline
 & Full Resolution (70 pixels/mm) & \multicolumn{1}{c|}{} & 0.109 & 0.883 & 4.960 & 0.208 & 0.865 & 0.949 & 0.975 \\ \hline \hline
 & Target Resolution (35 pixels/mm) & \multicolumn{1}{c|}{} & 0.117 & 1.001 & 5.144 & 0.213 & \textbf{0.855} & 0.946 & 0.974 \\ \cline{2-10} 
 & Wide Angle Camera (30 pixels/mm) & \multicolumn{1}{c|}{} & 0.119 & 1.026 & 5.202 & 0.216 & 0.850 & 0.943 & 0.974 \\ \cline{2-10} 
 & \textbf{Ours no weighting} & \multicolumn{1}{c|}{} & \textbf{0.115} & \textbf{0.942} & 5.087 & 0.209 & 0.853 & \textbf{0.948} & 0.976 \\ \cline{2-10} 
 & \textbf{Ours fovea weighting} & \multicolumn{1}{c|}{\chk} & 0.116 & 0.950 & \textbf{5.038} & \textbf{0.206} & 0.852 & \textbf{0.948} & \textbf{0.977} \\ \cline{2-10} 
 & Color edges no weighting & \multicolumn{1}{c|}{} & 0.122 & 0.974 & 5.278 & 0.220 & 0.836 & 0.940 & 0.973 \\ \cline{2-10} 
\multirow{-6}{*}{(a)} & Color edges fovea weighting & \multicolumn{1}{c|}{\chk} & 0.123 & 0.958 & 5.267 & 0.220 & 0.831 & 0.940 & 0.974 \\ \hline \hline
 & Target Resolution (27 pixels/mm) &  & \textbf{0.118} & 1.013 & 5.209 & 0.215 & \textbf{0.848} & 0.943 & 0.974 \\ \cline{2-10} 
 & Wide Angle Camera (23 pixels/mm) &  & 0.121 & 0.996 & 5.264 & 0.219 & 0.839 & 0.940 & 0.973 \\ \cline{2-10} 
 & \textbf{Ours no weighting} &  & 0.121 & 1.003 & 5.192 & \textbf{0.211} & 0.844 & \textbf{0.945} & \textbf{0.976} \\ \cline{2-10} 
 & \textbf{Ours fovea weighting} & \multicolumn{1}{c|}{\chk} & 0.119 & \textbf{0.938} & \textbf{5.161} & \textbf{0.211} & 0.842 & 0.944 & \textbf{0.976} \\ \cline{2-10} 
 & Color edges no weighting &  & 0.137 & 1.124 & 5.721 & 0.247 & 0.797 & 0.920 & 0.964 \\ \cline{2-10} 
\multirow{-6}{*}{(b)} & Color edges fovea weighting & \multicolumn{1}{c|}{\chk} & 0.134 & 1.056 & 5.660 & 0.240 & 0.801 & 0.924 & 0.967 \\ \hline \hline
 & Target Resolution (8 pixels/mm) &  & 0.194 & 2.705 & 7.378 & 0.296 & 0.730 & 0.889 & 0.949 \\ \cline{2-10} 
 & Wide Angle Camera (7 pixels/mm) &  & 0.234 & 4.144 & 8.317 & 0.330 & 0.686 & 0.867 & 0.937 \\ \cline{2-10} 
 & \textbf{Ours no weighting} &  & 0.167 & 1.516 & 6.815 & 0.270 & 0.743 & 0.900 & 0.958 \\ \cline{2-10} 
 & \textbf{Ours fovea weighting} & \multicolumn{1}{c|}{\chk} & \textbf{0.164} & \textbf{1.463} & \textbf{6.555} & \textbf{0.256} & \textbf{0.754} & \textbf{0.909} & \textbf{0.964} \\ \cline{2-10} 
 & Color edges no weighting &  & 0.167 & 1.514 & 6.836 & 0.273 & 0.741 & 0.898 & 0.957 \\ \cline{2-10} 
\multirow{-6}{*}{(c)} & Color edges fovea weighting & \multicolumn{1}{c|}{\chk} & 0.167 & 1.472 & 6.589 & 0.260 & 0.747 & 0.907 & 0.963 \\ \hline
\end{tabular}}
\caption{SaccadeCam compared against equiangular (conventional) images. For a variety of bandwidth ratios of full resolution vs. target resolution, we show how the SaccadeCam framework (shown in Fig. \ref{fig:method}) outperforms target resolution images with conventionally uniformly distributed resolution across the FOV.
}
\label{tab:abl1}
\end{table*}

%% file: sections/hardware.tex
\section{SaccadeCam Hardware Prototype}
\label{sec:hardware}

Here we discuss a physical instantiation of SaccadeCam that can adaptively distribute resolution onto regions of interest based on our trained models. SaccadeCam consists of a low-resolution wide angle camera (WAC) whose field-of-view (FOV) covers the scene, and a narrow FOV telephoto camera that views reflections off a small, fast moving microelectromechanical (MEMS) mirror. These components are collectively the SaccadeCam device seen in Fig. \ref{fig:teaser}. 

Unlike many other MEMS mirror enabled devices (such as LIDARs ~\cite{flatley2015spacecube,stann2014integration,krastev2013mems}), we do not run our MEMS mirror at resonance. Instead we use a specific scan pattern, and we are able to control 5 points (i.e. 5 fovea) in the FOV at 5 Hz. This speed is reasonably fast for most objects in common scenes for depth inference. Our telephoto and WAC cameras consist of a 1.6 MP FLIR Blackfly S-U3-16S2C-CS, where the telephoto camera has a 30mm lens and the WAC camera has a 6mm lens. The telephoto camera views reflections off a 3.6mm Mirrorcle Technologies MEMS mirror with custom modifications to prevent ghosting artifacts induced from MEMS electronic packaging. Our main computer is a NVIDIA Jetson NANO, a popular embedded board with GPU and CUDA capabilities. We trace our PyTorch models to TorchScript so we can run our models on-device in C++. The Jetson NANO communicates with custom synchronization circuitry containing a Teensy 4.0 microcontroller that triggers the cameras and MEMS mirror in lockstep. The MEMS mirror is physically controlled from the Teensy through a Mirrorcle Technologies PicoAmp 5.4 X200 Digital to Analog Converter. Our hardware prototype is capable of on-device training although we leave this for future work. 

\subsection{Feasible Fovea from the Attention Mask}

In Sect. \ref{sec:algo} we discussed how to process the input, low-resolution WAC image to produce an attention mask across the WAC FOV, with the goal of increasing resolution in this region up to the bandwidth limit. Such an attention mask is deformable and non-convex, in the sense that there are no restrictions on optical feasibility of sensing the attention region in higher resolution, quickly.  

In this section we discuss how to extract a discrete number of optically feasible saccades from the attention mask for a practical MEMS-mirror-based SaccadeCam. We also contend that it will apply to any camera that is not capable of producing programmable spatially varying deformable point spread functions (PSFs). While phase masks \cite{wu2019phasecam3d} can achieve these types of deformable attention masks, they are both slow and work best with coherent light, rather than incoherent light from a scene. 

Our goal is  to maximize attention mask coverage with $n$ saccades, or mirror viewpoints. These correspond to $n$ pairs of voltages that specify the MEMS mirror viewpoints, $\{(\theta(V(t_1)), \phi(V(t_1))),...(\theta(V(t_n)), \phi(V(t_n)))\}$. We first tackle the problem of fixed foveal size or fovea FOV, and then we generalize such that each viewing direction $i$ could have its own unique FOV (perhaps using a liquid lens \cite{zohrabi2016wide}). 

\textbf{Greedy Attention Algorithm:} The greedy algorithm requires an attention mask and a \emph{fixed} angular fovea size $\omega_{fovea}$. Given an attention mask defined on the FOV, $\mathbf{A}(\omega)$ where $\omega \in \omega_{fov}$, we can find the location of the maximum attention value, $\omega_{max}$ in this mask. We then follow an iterative procedure, where we capture a fovea by selecting $t_1$ such that the first mirror direction $(\theta(V(t_1)), \phi(V(t_1)))$ points along the central axis of the solid angle defined by $\omega_{max}$. 
We then destroy attention mask information around the first maximum such that $\mathbf{A}(\omega)= 0$, where $\omega \in \omega_{fov} \ \ and \ \ \|\omega_{max} - \omega\| \leq \omega_{fovea}$. We then repeat the procedure $n$ times for $n$ fovea, until a set of mirror voltages are obtained $\{(\theta(V(t_1)), \phi(V(t_1))),...(\theta(V(t_n)), \phi(V(t_n)))\}$. 
The proof of this method follows from the greedy selection of subsequently maximum attention values, all of which are monotonically decreasing (i.e. $\omega_{max}$ for $t_1$ is less than $\omega_{max}$ at  $t_2$ and so on). Therefore, there is no way that there exists an attention value at location $\omega_{missed}$ that is greater than the $n$ selected values at different locations of $\omega_{max}$, because otherwise it would have been selected for measurement at some point between $t_1$ and $t_n$. We present derivations for an advanced attention coverage algorithm based on the optical knapsack algorithm from \cite{pittaluga2015privacy} in the supplementary, although we do not implement this algorithm in hardware.


\subsection{Hardware Prototype Results}

We show qualitative results on real data captured with our SaccadeCam hardware prototype in Fig. \ref{fig:hardware_results}. Our results are obtained on-device at video rate as follows. The NVIDIA Jetson NANO triggers the WAC camera and passes the WAC image through our trained attention network. Next, given a calibrated MEMS mirror, telephoto and WAC cameras, we determine the top ten pixel locations (and therefore MEMS voltages) that optimally cover the attention prediction with our greedy algorithm. The mirror is triggered and moves to a location whereby the telephoto camera is subsequently triggered to capture an image of the MEMS mirror reflection. We choose ten fovea for our hardware prototype so the previous step is repeated until ten MEMS mirror images are captured; the Hardware Attention column in Fig. \ref{fig:hardware_results} shows examples of the ten captured MEMS mirror images taken by the telephoto camera. We then gamma correct and blend the telephoto camera images onto the WAC image to form the SaccadeCam image. Finally, the SaccadeCam image is passed through our depth network to obtain our result. 

For results with our SaccadeCam hardware prototype, we keep the target resolution bandwidth at 35 px/mm and SaccadeCam WAC bandwidth at 31 px/mm with ten fovea. This lets us use models trained on the much larger KITTI dataset. For target depth we use the 20 epoch weights of 35 px/mm target bandwidth. We finetune SaccadeCam weights for 5 epochs at 1e-7 learning rate on KITTI with patch fovea to smooth out rough square boundary edges occurring when overlaying fovea images onto the WAC image since the fovea images are square and do not perfectly approximate the learned attention. 

Fig. \ref{fig:hardware_results} shows that our hardware prototype can qualitatively match the results seen on the KITTI test set in Fig. \ref{fig:results} in that SaccadeCam depth outperforms target depth in the learned attention regions thanks to the natively-high angular resolution of the telephoto camera viewing the MEMS mirror.